\documentclass[letterpaper,10pt,conference]{ieeeconf}

\IEEEoverridecommandlockouts
\usepackage{graphicx}
\usepackage{amsmath}
\usepackage{amssymb}
\usepackage{amsthm}
\usepackage{color}
\usepackage{subcaption}

\newcommand{\eg}{\textit{e.g.}}

\newcommand{\typeone}{\textsc{Type I\ }}
\newcommand{\typetwo}{\textsc{Type II\ }}
\newcommand{\typethree}{\textsc{Type III\ }}
\newcommand{\RN}[1]{%
  \textup{\uppercase\expandafter{\romannumeral#1}}%
}

\newtheorem{problem}{Problem}
\newtheorem{theorem}{Theorem}

\begin{document}

\title{\LARGE \bf Algorithms for Routing of Unmanned Aerial Vehicles\\ with Mobile Recharging Stations}

\author{Kevin Yu, Ashish Kumar Budhiraja, and Pratap Tokekar%
\thanks{The authors are with the Department of Electrical \& Computer Engineering, Virginia Tech, U.S.A. Email: \texttt{\small \{klyu, ashishkb, tokekar\}@vt.edu}}%
\thanks{This material is based upon work supported by the National Science Foundation under Grant No. 156624 and NIFA grant 2015-67021-23857.}}

\maketitle

\begin{abstract}
We study the problem of planning a tour for an energy-limited Unmanned Aerial Vehicle (UAV) to visit a set of sites in the least amount of time. We envision scenarios where the UAV can be recharged along the way either by landing on stationary recharging stations or on Unmanned Ground Vehicles (UGVs) acting as mobile recharging stations. This leads to a new variant of the Traveling Salesperson Problem (TSP) with mobile recharging stations. We present an algorithm that finds not only the order in which to visit the sites but also when and where to land on the charging stations to recharge. Our algorithm plans tours for the UGVs as well as determines best locations to place stationary charging stations. While the problems we study are NP-Hard, we present a practical solution using Generalized TSP that finds the optimal solution. If the UGVs are slower, the algorithm also finds the minimum number of UGVs required to support the UAV mission such that the UAV is not required to wait for the UGV. Our simulation results show that the running time is acceptable for reasonably sized instances in practice.
\end{abstract}

\IEEEpeerreviewmaketitle

\section{Introduction}
Unmanned Aerial Vehicles (UAVs) are being increasingly used for applications such as surveillance~\cite{michael2011persistent}, package delivery~\cite{thiels2015use}, infrastructure inspection~\cite{liu2014review,ozaslaninspection}, environmental monitoring~\cite{dunbabin2012environmental}, and precision agriculture~\cite{das2015devices,tokekar2016sensor}. However, most small, multi-rotor UAVs have limited battery lifetime (typically $<$ 30 minutes) which prevents them from being used for long-term or large scale missions. There is significant work that is focused on extending the lifetimes of UAVs through new energy harvesting designs~\cite{morton2015solar}, automated battery swapping~\cite{toksoz2011automated}, low-level energy-efficient controllers~\cite{hosseini2013optimal}, and low-level path planning~\cite{morbidi2016minimum}. In this paper, we investigate the complementary aspect of high-level path planning with an emphasis on energy optimization.

\begin{figure}[ht]
\includegraphics[width=1\columnwidth]{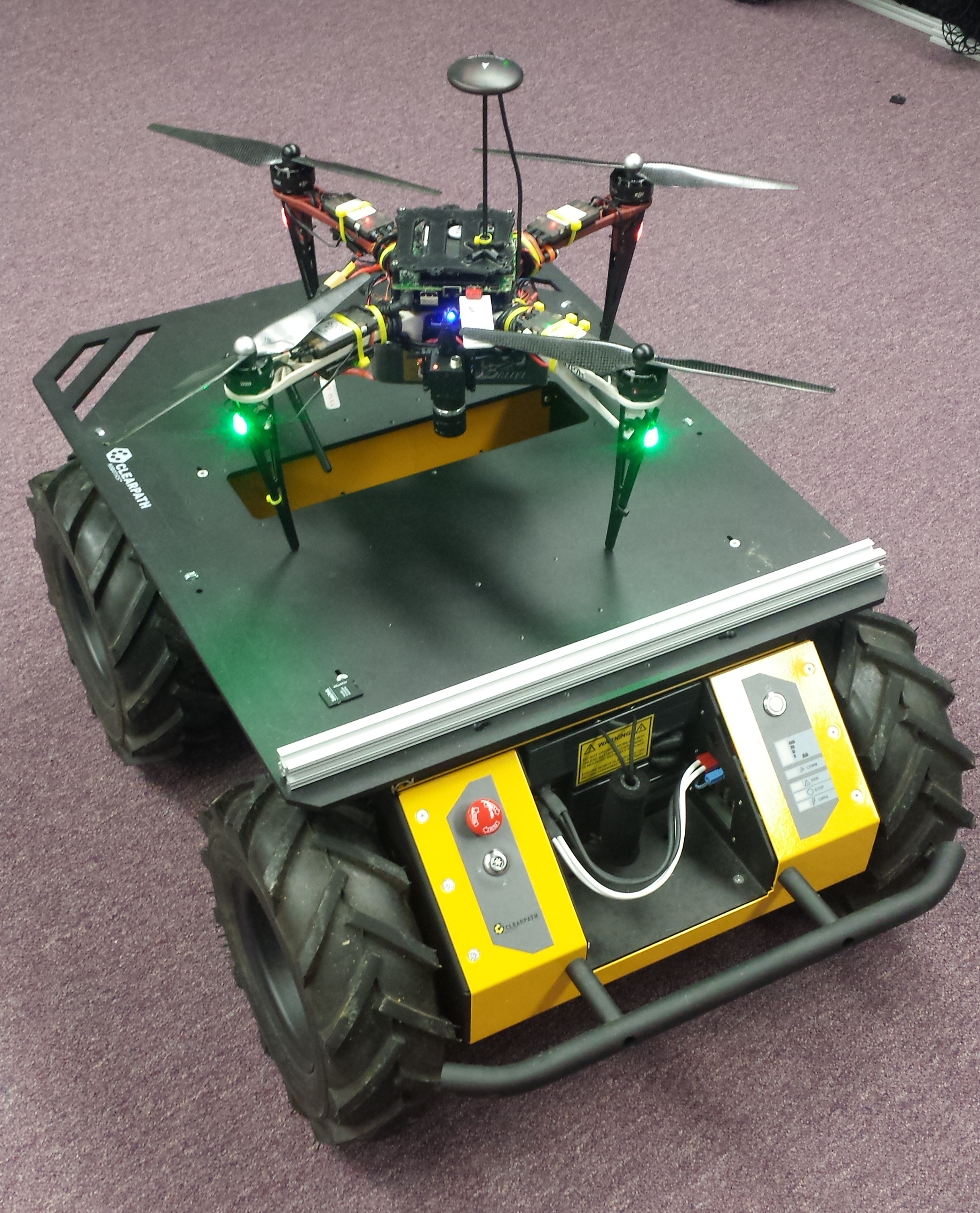}
\caption{DJI F450 with Pixhawk PX4 and NUC i7 on top of the Clearpath Husky used for the field experiments.}
\label{fig:realUAV}
\end{figure}

This work is motivated by persistent monitoring applications~\cite{smith2011persistent} where the UAVs are tasked with monitoring a finite set of sites on the ground by flying above these sites. The objective is to minimize the time required to visit all sites. In the absence of any additional constraints, this can be formulated as  Traveling Salesperson Problem (TSP) which is a classic optimization problem~\cite{arora1998polynomial}. However, when the sites are located far apart, the UAV may not have enough battery capacity to fly the entire tour. We consider scenarios where the UAVs are capable of landing on recharging stations and then taking off and continuing the mission. The recharging stations can either be stationary or placed on Unmanned Ground Vehicles (UGVs) which can charge the UAV while simultaneously transporting it from one site to another (Figure \ref{fig:realUAV}). This leads to a new variant of TSP where the output is not only a path for the UAV, but also a charging schedule that determines where and how much to recharge the UAV battery as well as paths for the UGVs. For a single UGV to keep up with the UAV, we also study the problem of minimizing the number of UGVs. Since it is not always possible for the UGV to keep up with the UAVs speed we implement a solver that can find the minimum number of UGVs necessary to service the UAV.

%A number of designs have been proposed for UAV recharging stations~\cite{mulgaonkar2014autonomous,cocchioni2014autonomous} including a commercial product that provides this capability out--of--the--box~\cite{SkySense.2016}. In this paper, we consider two types of problems: (i) finding the tour and the locations to deploy stationary recharging stations, and (ii) finding a tour for the UAV and one for an Unmanned Ground Vehicle (UGV) acting as a mobile basestation.
%The objective in both problems is to minimize the total time required to visit all sites. This includes the time to land and take-off from the recharging stations as well as the time to recharge the battery. 
%For the second problem, we also allow the UGV to recharge the robot while moving consequently saving time.

This problem generalizes Euclidean TSP ~\cite{arora1998polynomial} and is consequently NP-Hard. As such, assuming P$\neq$NP, no algorithm can guarantee the optimal solution in polynomial time. Instead, we seek algorithms that find the optimal solution in reasonable time for practical instances, similar to recent works~\cite{mathew2015multirobot,tokekar2016visibility}. Our main contribution is to show how to formulate both problems, described in section \ref{sec:probForm}, as Generalized TSP (GTSP)~\cite{noon1993efficient} instances. Earlier works have shown that a GTSP-based algorithm finds solutions faster than an Integer Programming approach~\cite{Smith2016GLNS,mathew2015multirobot,tokekar2016visibility}. We empirically evaluate two approaches to solve the GTSP instances: (1) GLNS solver~\cite{Smith2016GLNS} which uses heuristics to find potentially sub-optimal solutions in short time; and (2) an exact solver which reduces GTSP into TSP instances which are solved using concorde~\cite{applegate2006concorde}. We compare the time required to find the solution in both approaches. 

%Our problem formulation can also be used to solve a related problem of performing package deliveries at the sites with UAVs, where the UGVs are acting as mobile warehouses. The battery capacity constraint is replaced by a payload constraint. We briefly describe this and other modifications required.

\section{Related Work}
In this section we briefly describe the works related to UAV recharging stations.

\subsection{Recharging and Replacing UAV Batteries}
A number of solutions for autonomous charging of UAVs have been proposed in the recent past. Cocchioni et al.~\cite{cocchioni2014autonomous} presented a vision system to align the UAV with a stationary charging station. % They described hardware that allows for smooth and accurate alignment of a UAV's contact points to the stationary basestation's recharging contact points.
%The paper claims that 95\% of the time a UAV will land within the charging station. 
A similar design was presented by Mulgaonkar and Kumar~\cite{mulgaonkar2014autonomous} which included magnetic contact points. There are also commercial products (\eg, the SkySense system \cite{SkySense.2016}) that provide similar capabilities.

The alternative to recharging batteries is to swap them. Swieringa et al.~\cite{swieringa2010autonomous} presented a ``cold'' swap system for exchanging the batteries for one or more helicopters. The authors evaluated their system through simulations with three helicopters where they demonstrated an increase in system lifetime from six minutes to thirty two minutes. % The reason that this system was not able to obtain persistent operation is because they ran out of new batteries to swap.  In theory if there is enough batteries to swap then the system could obtain persistent operations. 
Toksoz et al.~\cite{toksoz2011automated} presented the design of a stationary battery swapping station for multi-rotor systems. Their design has a ``dual-drum structure'' that can hold a maximum of eight batteries which can be ``hot'' swapped. 
%By being able to ``hot" swap batteries, the system does not need to fully shut down to replace the battery. This gives a fully autonomous system the ability to run missions such as persistent surveillance. 

The work presented in this paper is complementary to these hardware designs --- any of the existing systems could be leveraged. Instead we show how to optimize the performance by careful placement of charging stations or planning of paths for mobile charging stations.

% In this paper we look into using mobile recharging stations. Stationary recharging stations allow for persistent missions and monitoring of UAV health however we would like to expand this to mobile recharging stations. We have created an algorithm that has the ability to handle ``hot'' and ``cold'' swappable battery systems. Overall by understanding how a stationary recharging basestations work we are able to create a better system for multi-robot persistent graph exploration. So far stationary basestations for recharging or battery swapping has been studied, but the new problem of placing the recharging station on a mobile robot presents other problems.

\subsection{Planning for Energy Limited UAVs} \label{sec:planningEnergy}
A typical strategy to deal with limited battery life of UAVs is to use multiple robots with possibly redundancy built in. Derenick et al.~\cite{derenick2011energy} presented a control strategy to carry out persistent coverage missions with robot teams which balances a weighted sum of mission performance and the safety of the UAVs. % The UAVs reconfigure based on their energy levels and coverage performance.
Mitchell et al.~\cite{mitchell2016persistent} presented an online approach for maintaining formations while  substituting UAVs running low on charge with recharged UAVs. Liu and Michael~\cite{liu2014energy} presented a matching algorithm for assigning UAVs with UGVs acting as recharging stations. %In this work we focus on planning with a single UAV with occasional recharging.

In our previous work~\cite{tokekar2016sensor}, we showed how to plan tours for a symbiotic UAV+UGV where the UGV can mule the UAV between two deployment locations such that the UAV does not spend any energy. However, the previous work did not model the capability of UGV recharging the UAV along the way. Consequently, the goal was to maximize the number of sites that can be visited in a single charge. This results in a variant of an NP-Hard problem known as orienteering~\cite{blum2007approximation}. In the current work, we allow for a more general model which has the added complication of keeping track of the energy level of the UAV as well as deciding where and how much to charge along the tour.
	
The work by Rathinam et al.~\cite{sundar2016formulations} looks into how to plan paths for mobile charging stations. The author of this paper studies the problem of planning UAV paths with fuel constraints and stationary refueling locations. The algorithms that the author presents allow for the planning of multiple UAVs to visit every site. This paper is slightly different than ours in the sense that the refueling stations are stationary and the UAV has to adjust to the refueling station. In our paper we allow the refueling station to mobile and adjust for the UAV allowing for shorter tour times.

%Moridian et al.~\cite{moridian2017postdisaster} presented a system of autonomous robots that can link together to allow for mobile power distribution. They use a ``source'' robot to obtain power and then ``bus'' and ``cable'' robots to distribute the power to disaster areas. We consider a simpler scenario of a single UAV and UGV but focus on the planning aspects of the problem.

The work most closely related to ours is that of Maini and Sujit~\cite{maini2015cooperation}. They present an algorithm that plans paths for one UAV and one recharging UGV to carry out surveillance in an area. The UGV moves on a road network. % They then locate a set of points that have to be visited by the UAV and create an circular area around them based on the maximum distance a UAV is able to travel while having the ability to return to the UGV which remains stationary.
The authors create an initial path for the UGV and then create a path for the UAV. In this paper, we simultaneously create paths for the UAV and UGV. Additionally, we guarantee that our algorithm finds the optimal solution for the problem.

\section{Problem Formulation} \label{sec:probForm}
In this section we formally define the problem. Throughout the paper, we focus on the main problems of planning with limited battery lifetime.% We briefly note how the results can be extended to planning for package delivery. 

The input to our algorithm is a set of $n$ sites, $x_i$, that must be visited by the UAV. We start with a list of common assumptions:
\begin{enumerate} \label{assumptions}
\item unit rate of discharge (1\% per second);
%\item all vertex/edges can be traversed by either thUGV, UAV, or UGV+UAV;
\item UAV has an initial battery charge of 100\%;
\item UAV and UGVs start at a common \emph{depot}, $d$;
\item all the sites are at the same altitude;
\item UAV can fly between any two sites if it starts at 100\% battery level;
\item UGVs have unlimited fuel/battery capacity.
% \item The UGV incurs no cost penalty, only the UAV incurs a cost penalty
% \item Once the UAV lands on the UGV it will be instantaneously fully charged
% \item $C_{it} + C_{ti} \le 100\%$, This states the cost of traveling to and from a point has to be less than or equal to $100\%$ of battery life
\end{enumerate}
All but the last assumption are only for the sake of convenience and ease of presentation and can be easily relaxed. Although UGVs cannot have unlimited operational time, it is a reasonable assumption since UGVs can have much larger batteries or can be refueled quickly.
%, if using an internal combustion engine, relative to the UAV.

We also provide a list of standard terminology that will be used throughout this paper:
\begin{itemize} \label{terminology}
\item $x_i$ denotes the $i^{th}$ site that must be visited\footnote{Note that $x_i$ does not mean that is the $i^{th}$ point that will be visited. The order of visiting the points is determined by the algorithm.} by flying to a fixed altitude;
%\item $s_i$ denotes a charging station'' \pcomment{(only used if needed)}
%\item $g_i$ represents a \emph{cluster} of all possible $x_i^k$ at a site $x_i$, where $k$ corresponds to a given battery level;
\item $r$ represents the time required to recharge the battery by a unit \%;
\item $t_{TO}$ is the time it takes to take off from the UGV;
\item $t_{L}$ is the time it takes to land on the UGV;
\item $t_{UAV}(x_i,x_j)$ and $t_{UGV}(x_i,x_j)$ give the time taken by the UAV and UGV to travel from $x_i$ to $x_j$.
\end{itemize}

Suppose $\Pi$ is a path that visits the sites in the order given by $\sigma:
\{1,\ldots,n\}\rightarrow \{1,\ldots,n\}$ where $\sigma(j) = i$ implies $x_i$ is the $j^{th}$ point visited along $\Pi$. The cost of an edge from $x_{\sigma(j)}$ to $x_{\sigma(j+1)}$ along $\Pi$ depends on whether the UAV flies between the two sites or if it is muled by the UGV between the two sites while being recharged. Let $k$ and $k'$ be the battery levels at $\sigma(j)$ and $\sigma(j+1)$. Therefore, 
\begin{equation}
T(j,j+1) = \begin{cases}
t_{UAV}(x_{\sigma(j)},x_{\sigma(j+1)})\\
\max \{t_{UGV}(x_{\sigma(j)},x_{\sigma(j+1)}),r(k'-k)\}
\end{cases}
\end{equation}
In addition, we also have non-zero node costs if the UAV is charged from battery level $k$ to $k'$ at a site $x_i$ rather than along an edge:
\begin{equation}
T(j) = r(k'-k).
\end{equation}

Therefore, the total path cost is given by,
\begin{equation}
T(\Pi) = T(1) + \sum_{j=1}^{n-1} T(j+1) + T(j,j+1)
\label{eqn:pathcost}
\end{equation}

We are now ready to define the problems studied in this paper. 

\begin{problem}[Multiple Stationary Charging Stations (MSCS)] \label{problem1}
Given a set of sites, $x_i$, to be visited by the UAV, find a path $\Pi^*$ for the UAV that visits all the sites as well as select one or more sites (if needed) to place recharging stations so as to minimize the total time given by Equation~\ref{eqn:pathcost} under the assumptions given above.
\end{problem}

\begin{problem}[Single Mobile Charging Station (SMCS)] \label{problem2}
Given a set of sites, $x_i$, to be visited by the UAV, find a path $\Pi^*$ for the UAV that visits all the sites as well as another path for the UGV acting as a mobile basestation so as to minimize the total time given by Equation \ref{eqn:pathcost} under the assumptions given above. Assume that the UAV and UGV travel at the same speed.
\end{problem}

The assumption that the UGV is as fast as the UAV is not necessary to find a solution; it is required to guarantee optimality for one UGV. If the UGV is slower than the UAV, we can still use the paths returned by the algorithm for one UGV, but the UAV may have to wait. An alternative is to minimize the number of UGVs required to ensure the UAV never has to wait for a recharging station.

\begin{problem}[Multiple Mobile Charging Stations (MMCS)] \label{problem3}
Given a path, $\Pi^*$, for a UAV and a set of charging sites as well as \typetwo edges for the UGVs, find the minimum number of slower UGVs necessary to service the UAV, without the UAV having to wait for a UGV. The input to MMCS is obtained by solving SMCS, without assuming the single UGV in SMCS is as fast as the UAV.
\end{problem}

%\paragraph*{Package Delivery Variant} We can modify our proposed algorithm to address the package delivery problem. The recharging stations can be thought of as warehouses that load the UAV with the packages to be delivered and the battery level is the number of packages still onboard the UAV.
%The UAV has to return to a basestation when it runs out of packages to deliver and can carry multiple packages at the same time. The rest of the algorithm remains the same.

Our main contribution is a GTSP-based algorithm that solves the first two problems optimally and an Integer Linear Programming (ILP)-based algorithm that solves Problem \ref{problem3} optimally. As mentioned previously, Problems \ref{problem1} and \ref{problem2} are NP-Hard and consequently finding optimal algorithms with running time polynomial in $n$ is infeasible under standard assumptions. Instead, we provide a practical solution that is able to solve the three problems to optimality in reasonable time (quantified in Section~\ref{sec:sims}).

\section{GTSP-Based Algorithm}
In this section we show how to formulate Problems \ref{problem1} and \ref{problem2} as GTSP instances \cite{noon1993efficient}. The input to GTSP is a graph where the vertices are partitioned into clusters. The objective is to find a minimum cost tour that visits exactly one vertex per cluster. When each cluster contains only one vertex, the GTSP reduces to TSP. 

Solving GTSP is at least as hard as solving TSP. However, Noon and Bean \cite{noon1993efficient} presented a technique to convert any GTSP input instance into an equivalent TSP instance on a modified graph such that finding the optimal TSP tour in the modified graph yields the optimal GTSP tour in the original graph. We can solve GTSP by solving TSP optimally using a numerical solver and we use \emph{concorde} \cite{applegate2006concorde}, which is the state-of-the-art TSP solver or GLNS~\cite{Smith2016GLNS}, that is a heuristics-based GTSP solver. The results in Section~\ref{sec:sims} show that GLNS significantly faster than \emph{concorde}. However, only the \emph{concorde} approach is guaranteed to find the optimal solution.

We start by showing how to formulate the SMCS and MSCS problems as GTSP instances. After obtaining an output, we can convert the TSP solution back into a GTSP solution, then into a solution for the SMCS or MSCS problems. The process of converting SMCS and MSCS into TSP is the same. Only the process of converting the solution of TSP to solutions of SMCS and MSCS differ.

\subsection{Transforming SMCS/MSCS to GTSP}
\begin{figure}[ht]
\centering
\includegraphics[width=1\columnwidth]{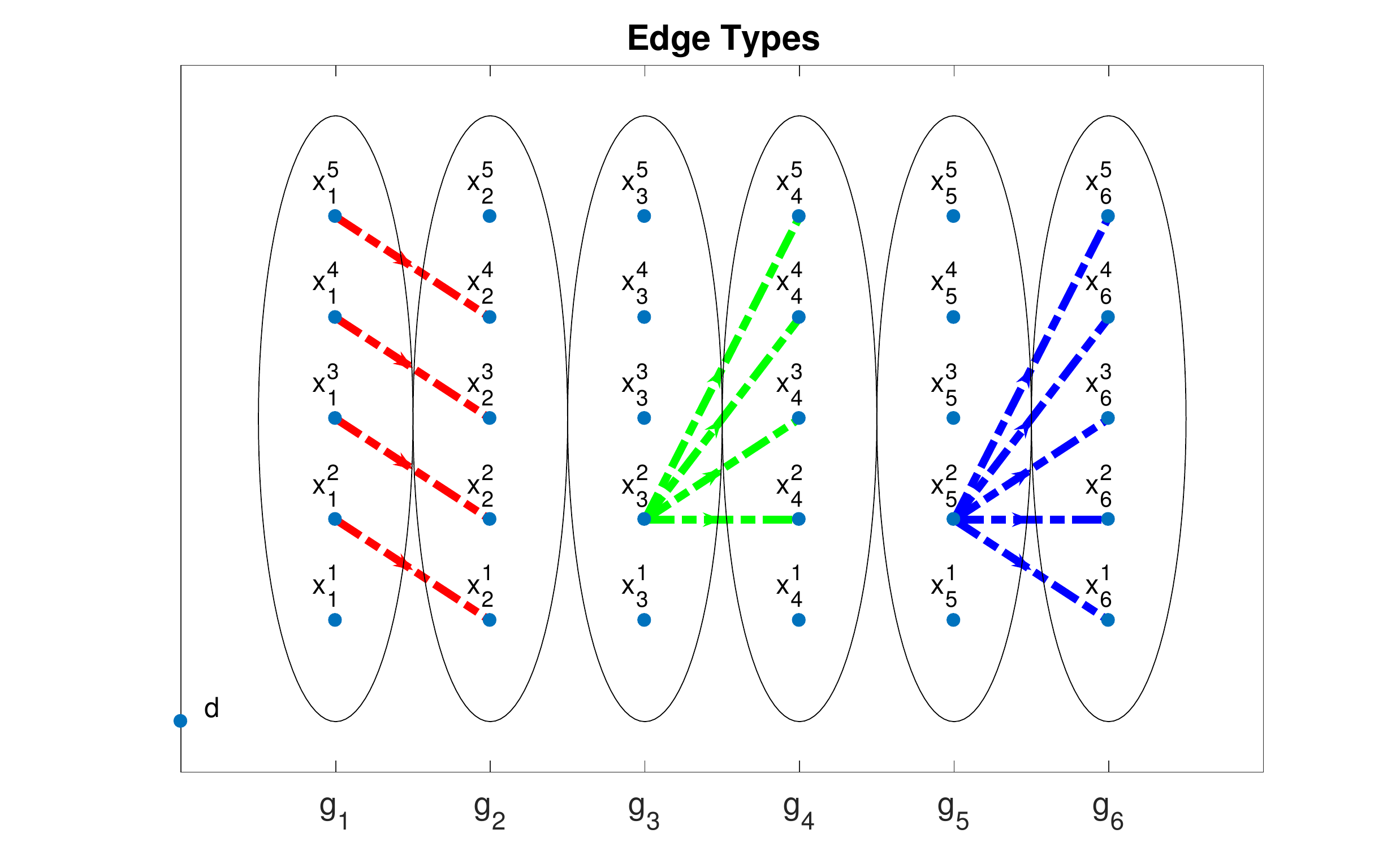}
\caption{Different types of edges that are created. \typeone is the red edges if there is a battery level drop of 1 between the two clusters, \typetwo is the green edge, and blue is \typethree. The depot station $d$ is shown on the graph. Note that only a subset of all possible edges are shown.\label{fig:edgetypes}}
\end{figure}

Given an SMCS or MSCS instance, we show how to create a GTSP instance consisting of a directed graph where the vertices are partitioned into non-overlapping clusters. We create one cluster, $g_i$, for each input site $x_i$. Each cluster, $g_i$ has $m$ vertices, each one corresponding to a discretized battery level. That is, $g_i = \{x_i^k \mid \forall i\in [1:n], \forall k\in \{1, 2, \ldots, m\}\}$. $x_i^k$ represents the UAV reaching site $x_i$ with $k * \frac{100\%}{m}$ battery remaining. $m$ is an input discretization parameter. Figure~\ref{fig:edgetypes} shows the six clusters for six input sites with $m=5$.

Next we describe how to create the edges amongst the vertices in the $n$ clusters. We create three types of edges. \typeone edge between $x_i^k$ and $x_j^{k'}$ models the case where the UAV directly flies from $x_i$ to $x_j$. The cost of a \typeone edge is given by:
\begin{equation*}
T_\text{I}(x_i^k,x_j^{k'}) = t_{UAV}(x_i, x_j)
\end{equation*}
A \typeone edge exists between $x_i^k$ and $x_j^{k'}$ if and only if $k-k'$ equals the distance between $x_i$ and $x_j$. For ease of exposition, we assume that taking-off and landing energy consumption is negligible. Nevertheless, we can easily incorporate this in the edge definitions. These types of edges are shown by the red lines in Figure \ref{fig:edgetypes}. 

A \typetwo edge from $x_i^k$ to $x_j^{k'}$ models the UAV landing on the UGV at $x_i$ and recharging while being muled to $x_j$ by the UGV. The cost of a \typetwo edge is given by:
\begin{equation*}
T_\text{II}(x_i^k,x_j^{k'}) = \max(r(k'-k), t_{UGV}(x_i, x_j)) + t_{TO} + t_{L}
\end{equation*}
The cost is the maximum of the time taken to recharge from $k$ to $k'$ and the time it takes the UGV to travel from $x_i$ to $x_j$. Note that a \typetwo edge exists only if $k' \geq k$. \typetwo edges are shown as the green edges in Figure \ref{fig:edgetypes}. 

Finally, we have \typethree edges that represent the UAV flying from $x_i$ to $x_j$ and then landing on the UGV at $x_j$ and recharging up to $k'$ battery level. The cost of a \typethree edge is given by:
\begin{equation*}
T_\text{III}(x_i^k,x_j^{k'}) = t_{UAV}(x_i, x_j) + r(k' - k + ||x_i-x_j||_2) + t_{TO} + t_{L}
\end{equation*}
A \typethree edge exists if and only if $k' \geq k - ||x_i-x_j||_2$. Figure~\ref{fig:edgetypes} shows the \typethree edges in blue. 

Only \typeone and \typethree edges exist when solving MSCS whereas all three edges are possible when solving MMCS. %\footnote{Since we assume that $t_{UAV}$ equals $t_{UGV}$, \typethree edges will never be part of the optimal tour. However, we present the most general version of the algorithm here.}
Note that \typetwo and \typethree edges require the UAV to take off and land at every site. This prevents the UAV from not taking off between two consecutive \typetwo edges. This is because in order to visit a site it must fly to a fixed altitude to consider the site visited.

There are certain pairs of vertices for which more than one type of edge may be allowed. In such a case, we pick the minimum of the three edge costs (assuming the edge cost is $\infty$ if the edge does not exist) and assign the minimum cost for the edge. That is, the edge cost $T(x_i^k,x_j^{k'})$ is given by:
\begin{equation*}
T(x_i^k,x_j^{k'}) = \min \{T_\text{I}(x_i^k,x_j^{k'}),T_\text{II}(x_i^k,x_j^{k'}),T_\text{III}(x_i^k,x_j^{k'})\}
\end{equation*}

We also create an $n+1^{th}$ cluster containing a dummy vertex called as the depot, $d$. We add a zero cost edge from $d$ to all vertices, $x_i^k$, with $k = m$ and edges from all vertices back to $d$. The reason to create a depot node is that the TSP solver finds a closed tour whereas we are interested in finding paths.\footnote{A path visits a vertex exactly once whereas a tour has the same starting and ending vertices.} The depot node serves to ensure that we can find a closed tour without charging for the extra edges.

%This graph with $n+1$ clusters represents the input GTSP instance. Using the method proposed by Noon and Bean~\cite{noon1993efficient}, we transform the GTSP instance into a TSP instance and solve for the optimal TSP tour using concorde~\cite{applegate2006concorde}.

\subsection{Converting Optimal TSP Tour to UAV and UGV Paths}
An optimal TSP tour immediately yields an optimal GTSP solution.
%An optimal GTSP solution will visit exactly one vertex in each cluster.
The order in which the clusters are visited gives the sequence of vertices on the UAV paths. What remains is deciding the UGV path for SMCS and recharging station placements for MSCS. 

In MSCS, we only have \typeone and \typethree edges. If a \typethree edge, say from $x_i^k$ to $x_j^{k'}$, appears in the GTSP solution, then we will place a recharging station at the site $x_j$. No recharging stations are placed for \typeone edges in the solution.

In MMCS, all three edges are possible, whereas only \typeone and \typetwo in SMCS. We check the type of each edge in the GTSP solution, one by one. If a \typeone edge appears in the GTSP solution, then it does not affect the UGV tour. If a \typetwo edge, say from $x_i^k$ to $x_j^{k'}$, appears in the GTSP solution, we add $x_i$ and $x_j$ to the UGV path (in this order). If a \typethree edge, say from $x_i^k$ to $x_j^{k'}$, appears in the GTSP solution we add only $x_j$ to the UGV path. The UGV path, as a result, visits a subset of the input sites. %Our algorithm only guarantees the optimal solution when the UAV and UGV have equal speeds.
If the UGV is slower than the UAV, then it is possible that the UAV will reach a site before the UGV does and will be forced to wait. We implement an ILP that allows us to solve for the minimum number of UGVs necessary to service the UAV without waiting (shown in section \ref{sec:ILP}).

\begin{theorem}
The GTSP-Based algorithm finds the optimal solution for SMCS and MSCS assuming that there exists an optimal TSP solution.
\end{theorem}
The proof follows directly from the proof of optimality of the GTSP reduction given by Noon and Bean~\cite{noon1993efficient}.% As described earlier, we use the concorde solver for finding optimal TSP solutions. We evaluate the computational times for solving practical instances in the next section.

\subsection{Solution for Problem \ref{problem3}} \label{sec:ILP}
We present a solution to the MMCS problem based on an ILP formulation. The input is obtained by solving Problem \ref{problem2}, where we are given a UAV path and a set of UGV sites and \typetwo edges. The UGV path visits only a subset of the sites in $\{x_i\}$. We denote these sites by $\{g_1, g_2, \ldots, g_l\}$, where $l\leq n$. For each edge from $g_i$ to $g_j$, where $(j>i)$, we associate the variable $y_{ij}$, which equals 1 if the edge will be traversed by some UGV and 0 otherwise. Also associated with each edge is the time to come for the UAV and UGV denoted by $T^{\Pi^*}(g_i,g_j)$ and $t_{UGV}(g_i,g_j)$ respectively. Here $T^{\Pi^*}(g_i,g_j)$ is the time taken by the UAV to fly the subpath of $\Pi^*$ from $g_i$ to $g_j$, which may contain some intermediate sites. $t_{UGV}(g_i,g_j)$, on the other hand is the time for the UGV to directly go from $g_i$ to $g_j$. Using the above notation we provide the ILP formulation given as follows:

\begin{equation}
\max{\sum_{i = 1}^{l-1}\sum_{j = i+1}^{l}y_{ij}}
\end{equation}

subject to:
\begin{equation}
\sum_{i=1}^{j-1}y_{ij}\le1 \text{  } \forall \text{  } j, \label{eq:2}
\end{equation}
\begin{equation}
\sum_{j=i+1}^{l}y_{ij}\le1 \text{  } \forall \text{  } j, \label{eq:3}
\end{equation}
\begin{equation}
y_{ij} = 0 \text{  if  } T^{\Pi^*}(g_i,g_j) < t_{UGV}(g_i,g_j), \text{  and}
\label{eq:4}
\end{equation}
\begin{equation}\label{eq:5}
y_{ij} = 1 \text{  if  } y_{ij} \text{  is \typetwo edge.}
\end{equation}

Equation \ref{eq:2} and \ref{eq:3} only allow a maximum of one incoming edge and a maximum of one outgoing edge. The constraint given by Equation \ref{eq:4} removes all UGV edges where the UAV would have to wait for the UGV. Lastly Equation \ref{eq:5} forces our problem to use \typetwo edges if present. Using the above equations we are able to solve Problem \ref{problem3}.

\section{Evaluations} \label{sec:sims}
In this section, we present simulation and preliminary experimental results using the proposed algorithm. 

\begin{figure*}[ht]
\centering
\begin{subfigure}[b]{0.3\textwidth}
\includegraphics[width=\textwidth]{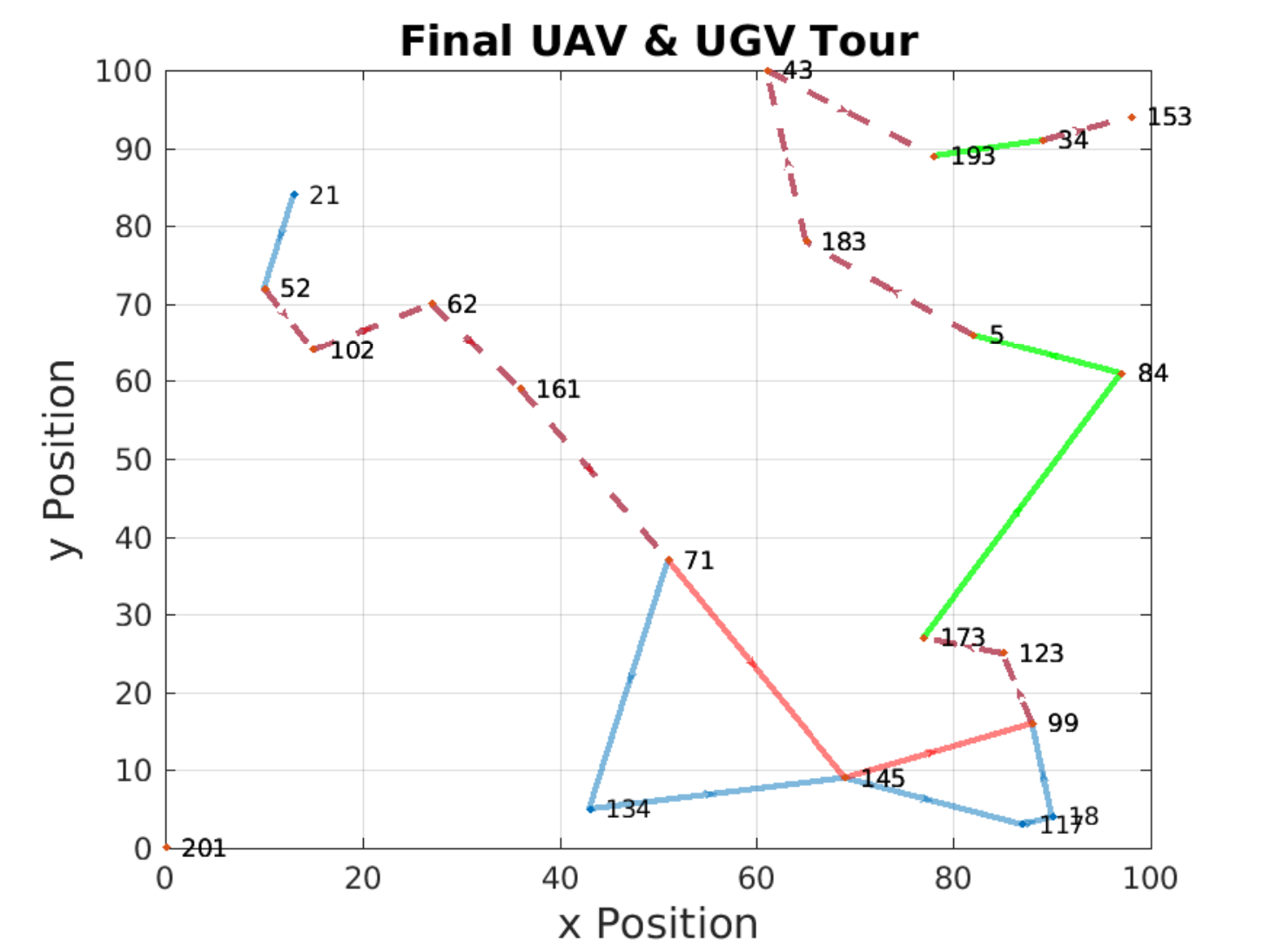}
\caption{UAV Tour and Cluster 0001 in order of visit.}
\label{fig:0001}
\end{subfigure}
\begin{subfigure}[b]{0.3\textwidth}
\includegraphics[width=\textwidth]{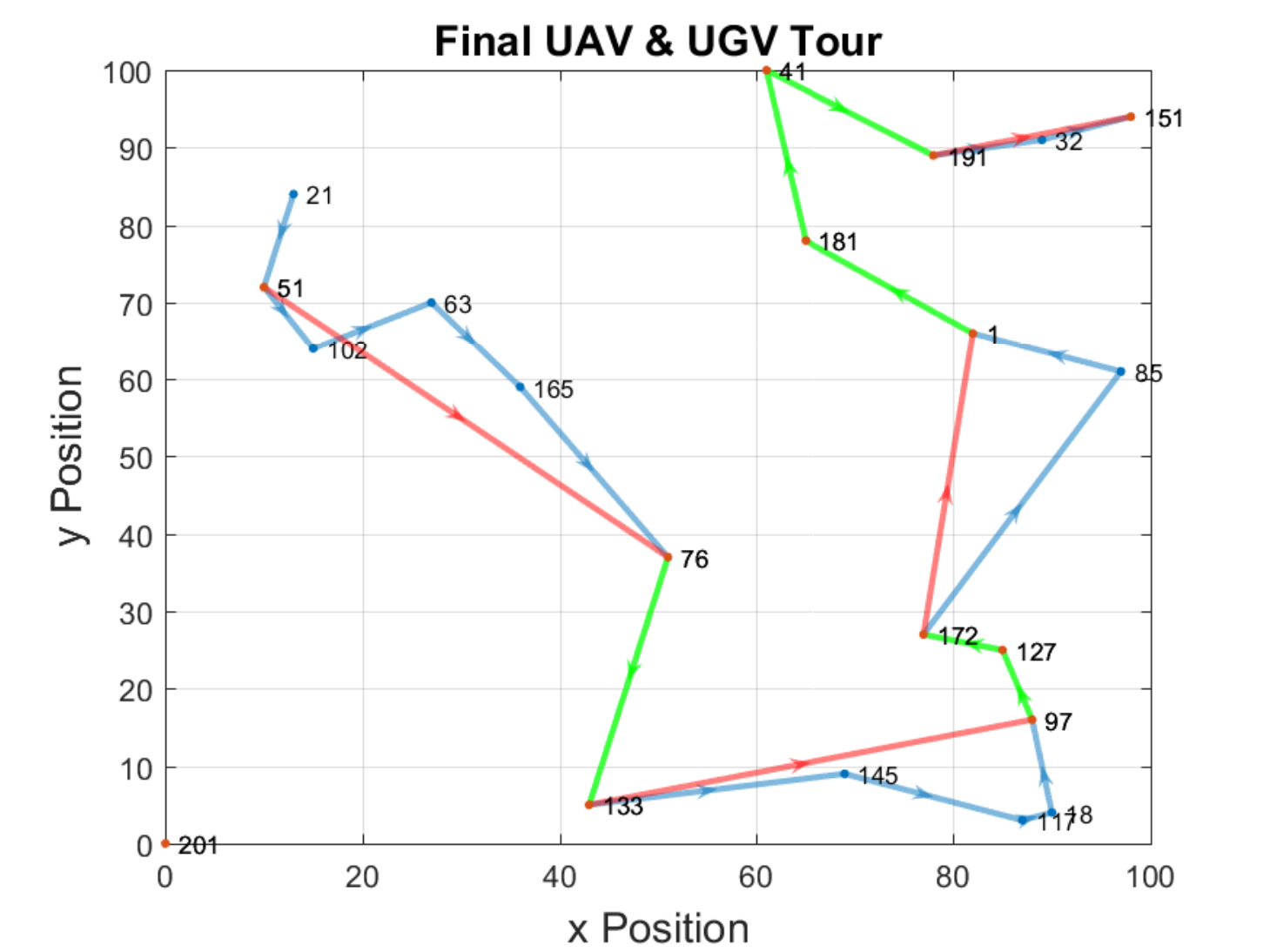}
\caption{UAV Tour and Cluster 0004 in order of visit.}
\label{fig:0004}
\end{subfigure}
\begin{subfigure}[b]{0.3\textwidth}
\includegraphics[width=\textwidth]{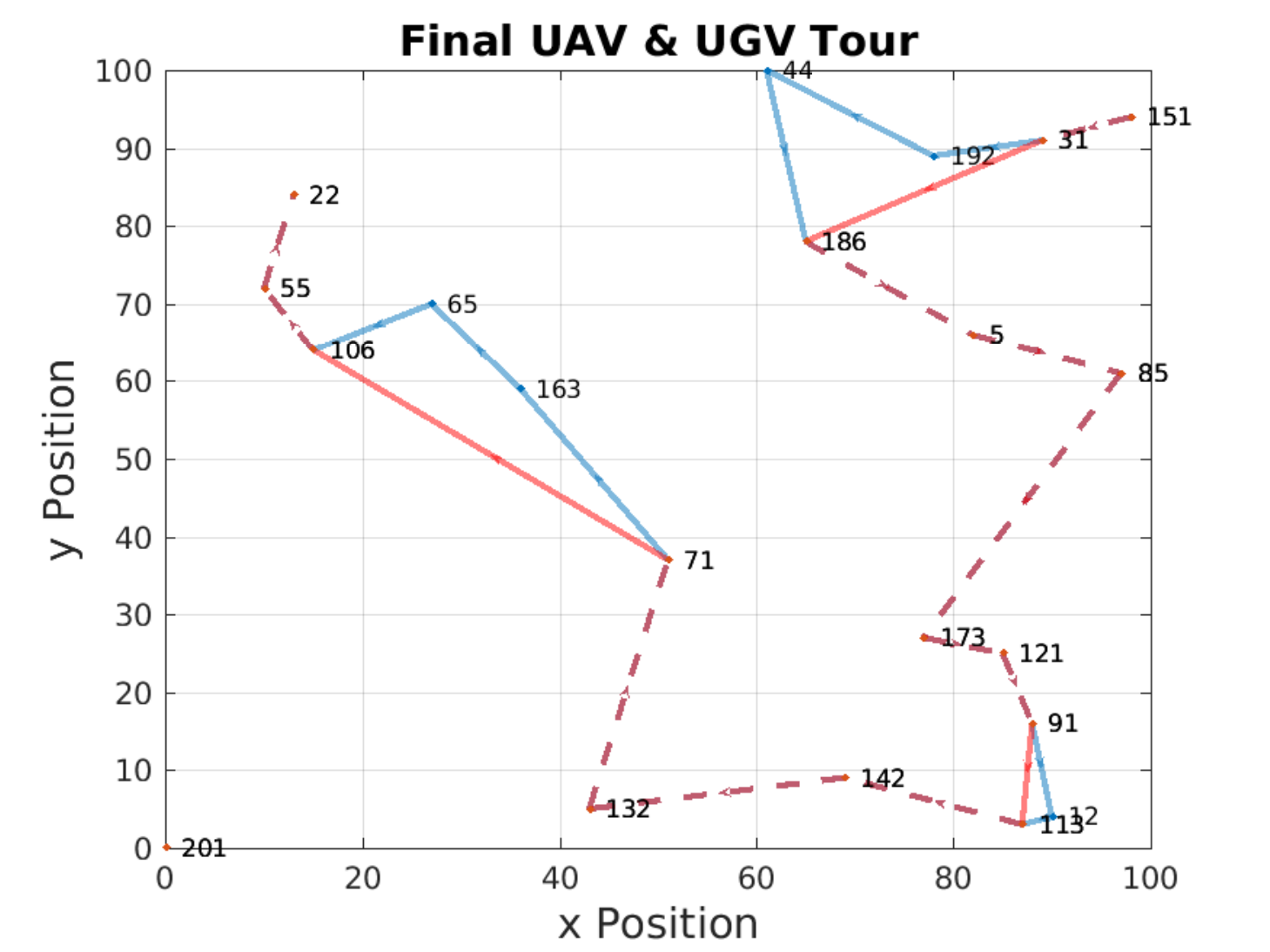}
\caption{UAV Tour and Cluster 0041 in order of visit.}
\label{fig:0041}
\end{subfigure}
\begin{subfigure}[b]{0.3\textwidth}
\includegraphics[width=\textwidth]{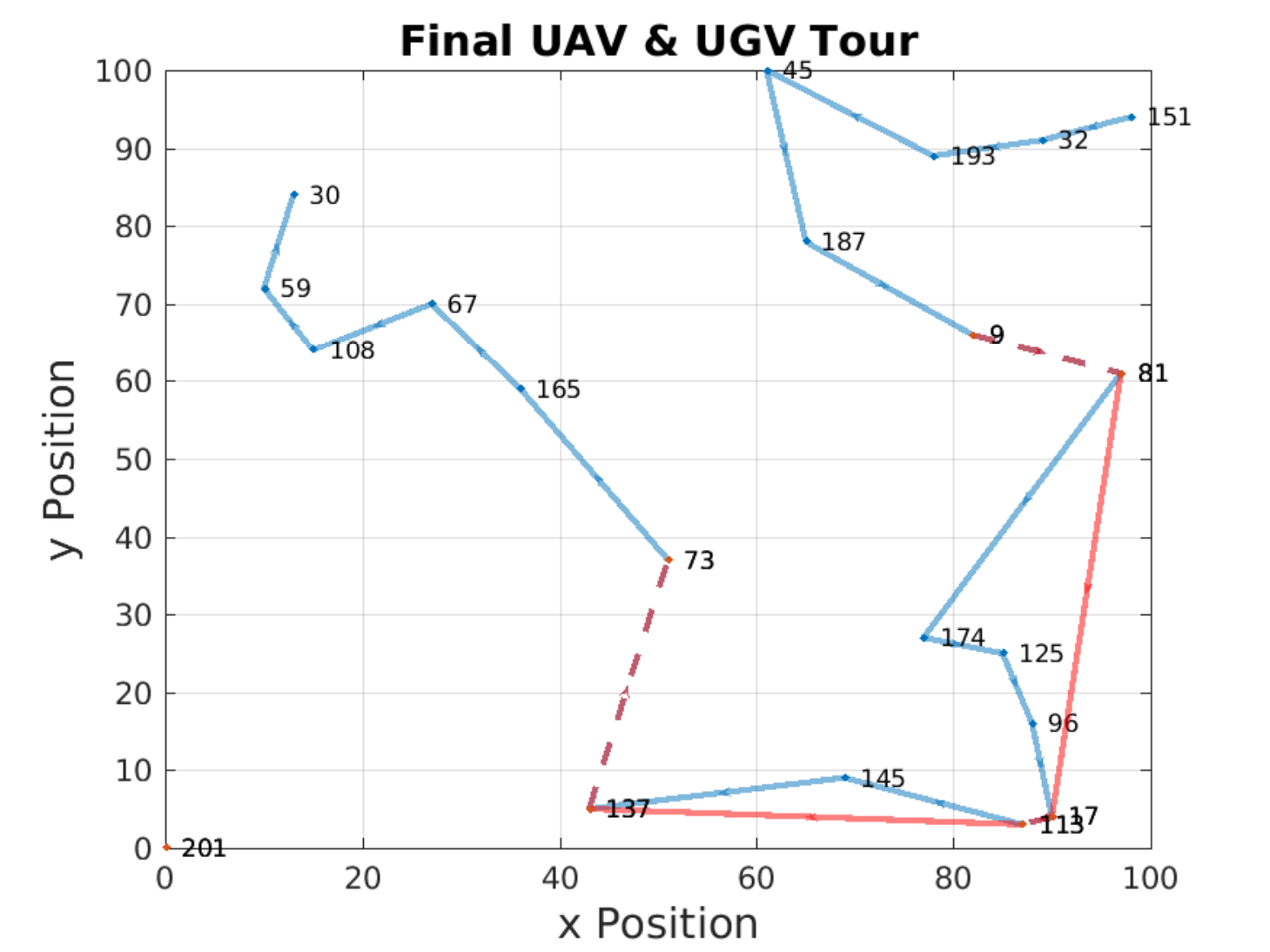}
\caption{UAV Tour and Cluster 4401 in order of visit.}
\label{fig:4401}
\end{subfigure}
\begin{subfigure}[b]{0.3\textwidth}
\includegraphics[width=\textwidth]{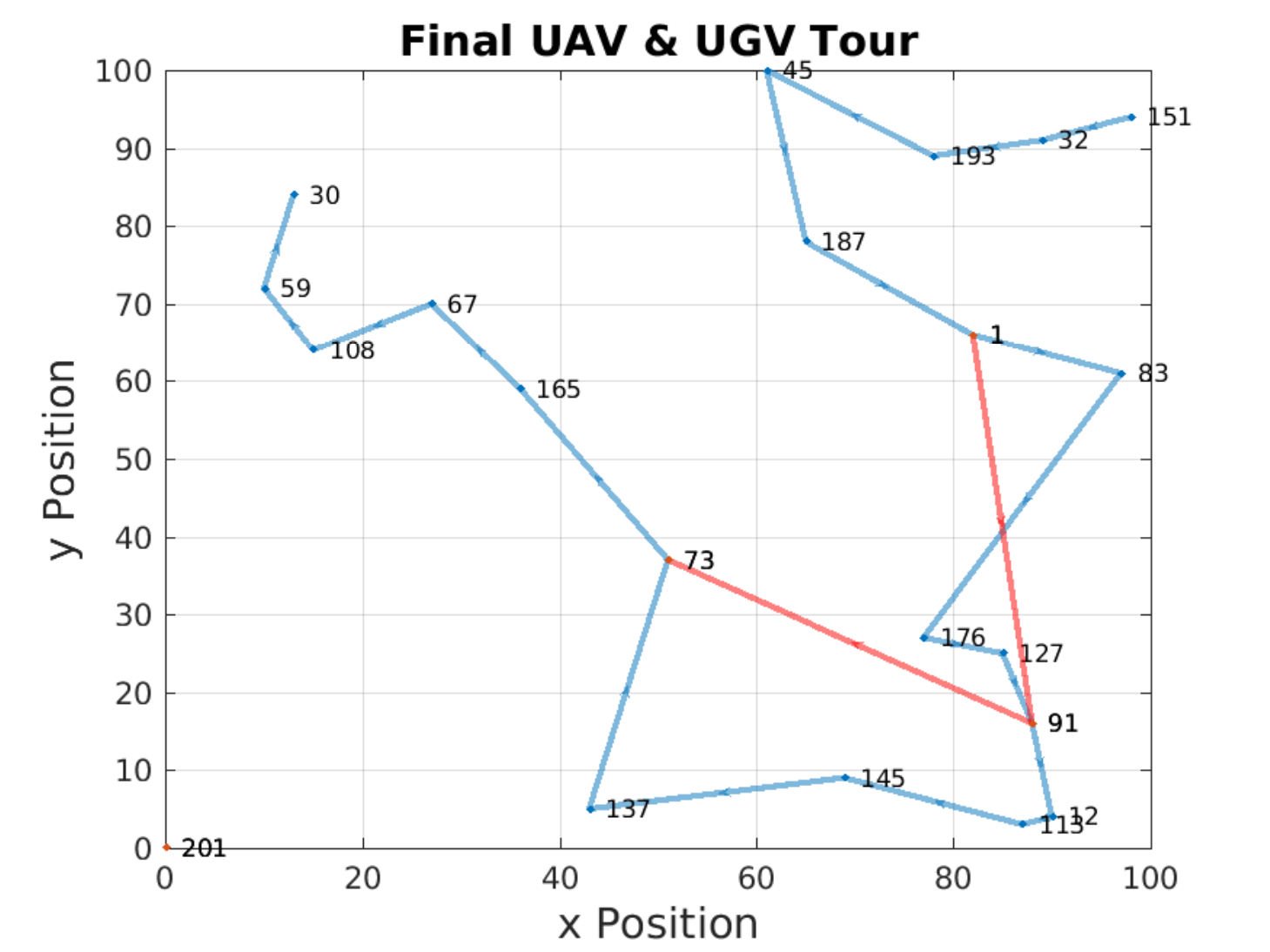}
\caption{UAV Tour and Cluster 4404 in order of visit.}
\label{fig:4404}
\end{subfigure}
\begin{subfigure}[b]{0.3\textwidth}
\includegraphics[width=\textwidth]{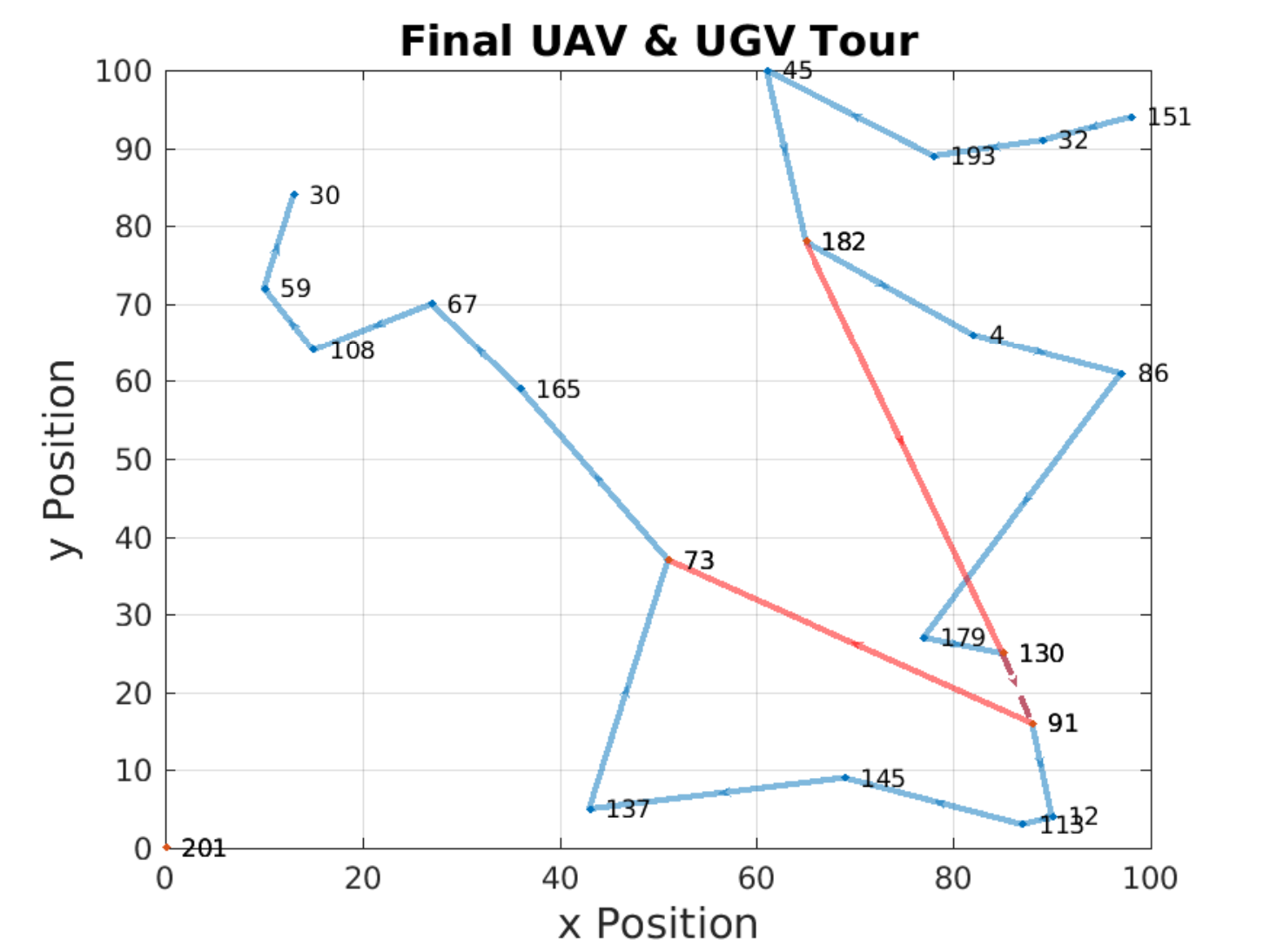}
\caption{UAV Tour and Cluster 4444 in order of visit.}
\label{fig:4444}
\end{subfigure}
\caption{The above figures are multiple runs using the same initial points, instead of randomizing them. We use 20 sites with 10 battery levels. Each image has the site number when in the GTSP formate to show battery level and a number ``WXYZ'' in the caption. This number ''WXYZ'' denotes: $t_{TO} = W$, $t_{L} = X$, $r = Y$, and $t_{UGV} = Z * t_{UAV}$. The colors represent different edge types with blue being only UAV travel, red being only UGV travel, green being UAV and UGV travel separate, and dashed red being UAV+UGV travel together. \label{fig:runs}}
\end{figure*}

\subsection{Effect of the Parameters} 

Figure~\ref{fig:runs} shows the outputs obtained for different configurations of the $t_{TO}, t_{L}, r, t_{UGV}$ parameters for the same 20 input sites and with $m=10$ battery levels. Each figure has the UAV+UGV tour with blue solid edges (only UAV), red solid edges (only UGV), green solid edges (UAV and UGV separate), and red dashed edges (UAV+UGV together). 

We make the following intuitive observations using the six cases shown in Figure~\ref{fig:runs}:
\begin{itemize}
\item $t_{TO} = 0$, $t_{L} = 0$ and $r = 0$: UAV does not differentiate between the type of the edge because there is no penalty to recharge (Figure~\ref{fig:0001});
\item $t_{TO} + t_{L} > 0$: recharging has a penalty and as such the number of recharging stops are reduced (Figures~\ref{fig:4401}, \ref{fig:4404} and \ref{fig:4444});
\item $t_{TO} = 0$, $t_{L} = 0$, $r = 0$ and $t_{UGV} > t_{UAV}$: the UAV will use \typethree edges for charging because $t_{UGV}$ will make \typetwo edges higher cost (Figure~\ref{fig:0004} and \ref{fig:4404});
\item $t_{TO} = 0$, $t_{L} = 0$, $r > 0$ and $t_{UGV} = t_{UAV}$: the UAV will use \typetwo edges for charging instead of \typethree edges (Figure~\ref{fig:0041});
\end{itemize}

We observe that %trends such as when $t_{TO} + t_{L} > 0$, the UAV will land for charging as few times as possible. 
the recharge time $r$ and UGV speed $t_{UGV}$ affect which type of edges are used. If the time it takes to recharge is much larger than $t_{UGV}$ then the UAV will favor  \typetwo edges and when the time it takes to recharge is much less than $t_{UGV}$ then the UAV will favor \typethree edges.% For all cases where $t_{UAV} = t_{UGV}$ the UAV will never have to wait for the UGV, but for cases where $t_{UAV} > t_{UGV}$ the UAV might have to wait at a site because the UGV is not fast enough to reach the site before the UAV. 

\subsection{Computational Time}
\begin{figure}[ht]
\centering
\begin{subfigure}[b]{0.46\columnwidth}
\includegraphics[width=\textwidth]{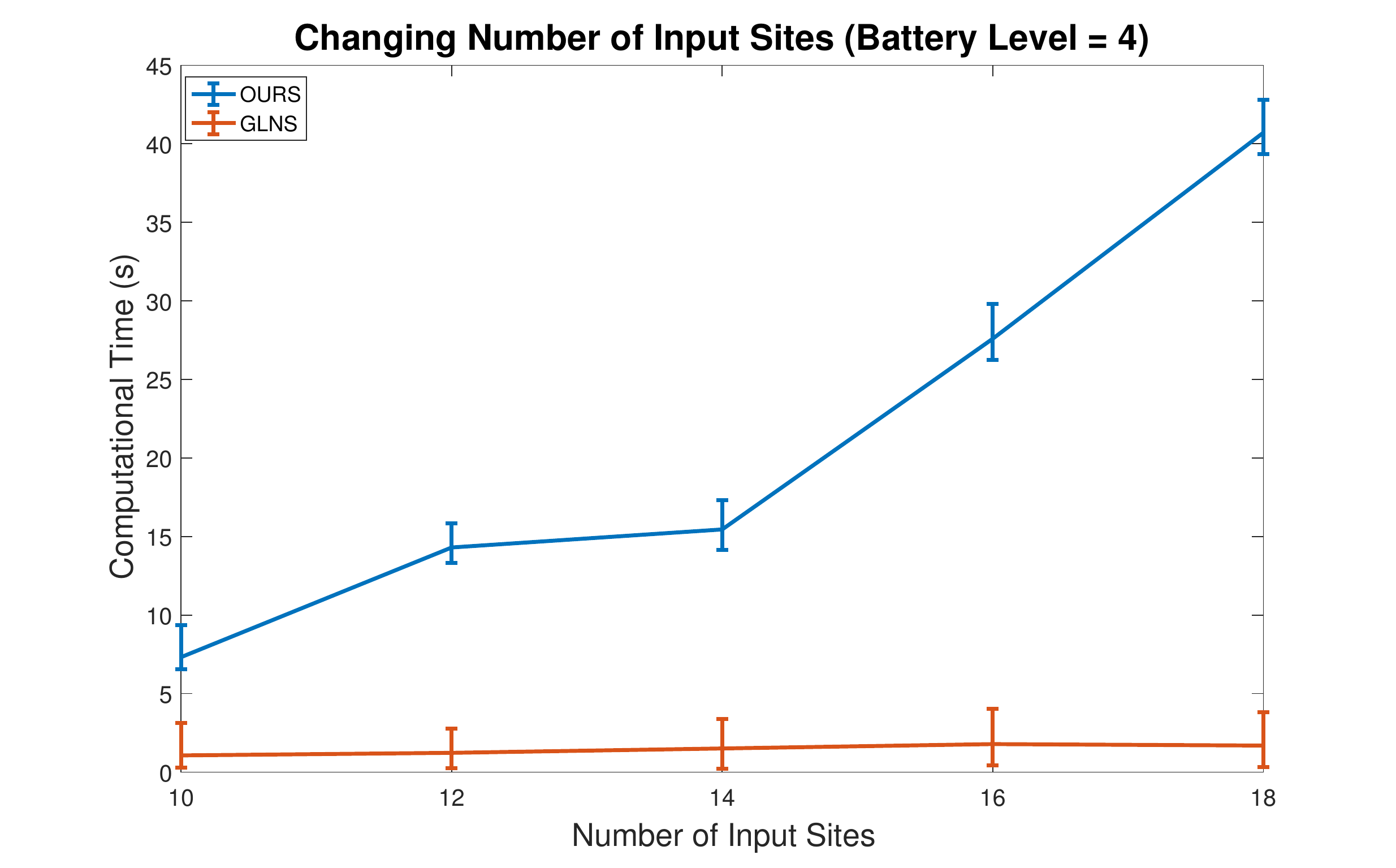}
\caption{Computational time with battery levels set to 4.}
\label{fig:sitechange}
\end{subfigure}
\begin{subfigure}[b]{0.45\columnwidth}
\includegraphics[width=\textwidth]{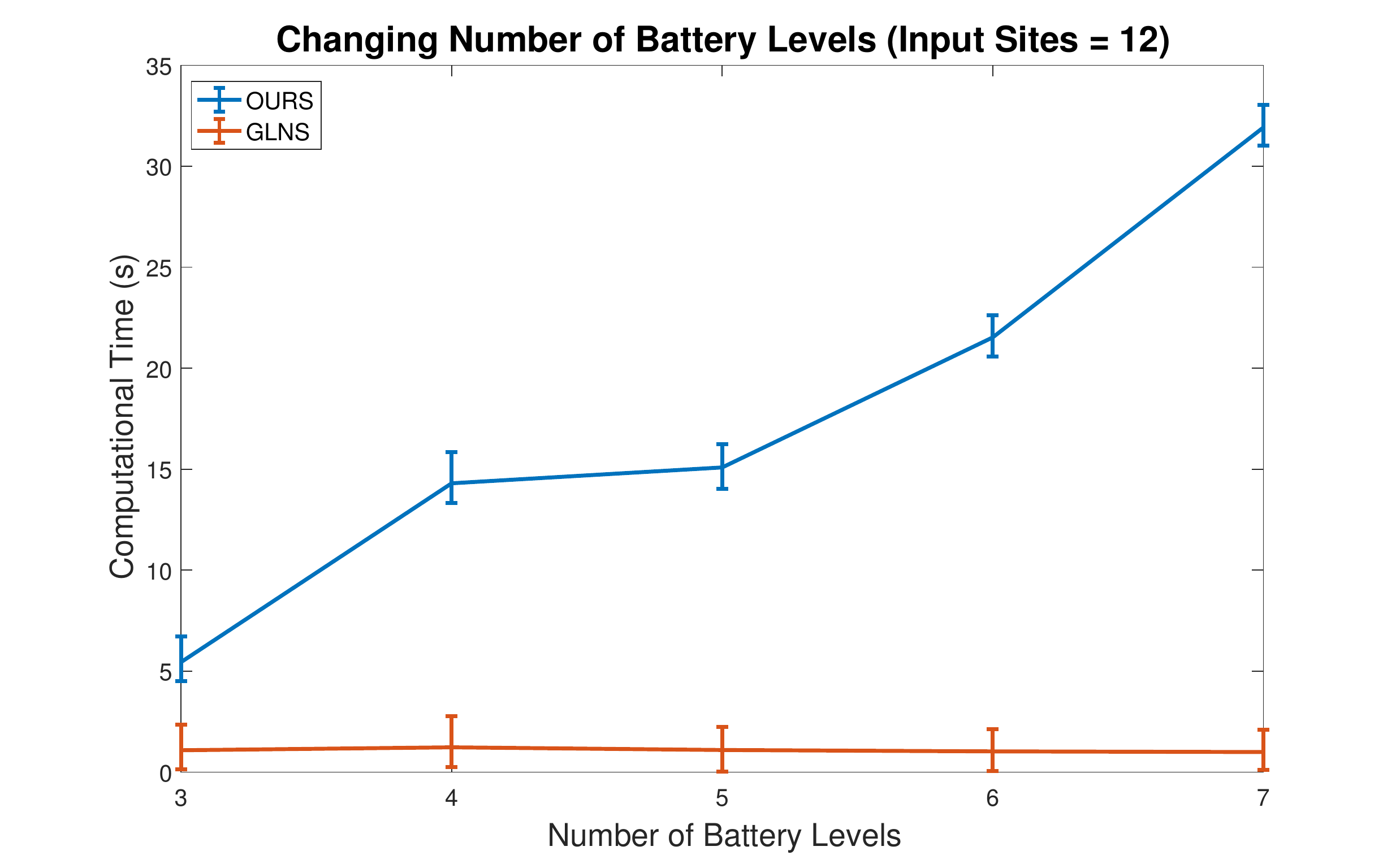}
\caption{Computational time with input sites set to 12.}
\label{fig:levelchange}
\end{subfigure}
\caption{Comparison of computational times of the GTSP to TSP transformation~\cite{noon1993efficient} solved using concorde and the direct GTSP solver, GLNS \cite{Smith2016GLNS} on the default ``medium'' setting. The output costs for the concorde and GLNS solutions were the same for the given instances, but may differ for larger instances.}
\label{fig:timecomparison}
\end{figure}

We use two solvers for the MSCS and SMCS problems. When using \emph{concorde} we obtain an optimal solution, but with more computational time. Therefore, it can only solve smaller instances. The GLNS solver can solve larger instances, but cannot always guarantee optimality. Nevertheless, we observe that the GLNS was able to find the optimal solution for all of the cases reported in Figure \ref{fig:timecomparison}.

Figure \ref{fig:timecomparison} shows a direct comparison of the computational times of the two methods. We compared the two methods by first varying the amount of inputs sites (Figure \ref{fig:sitechange}), with $m = 4$, and then comparing the two by varying the amount of battery levels (Figure \ref{fig:levelchange}), given $n = 12$. We ran 10 trials with random input sites. We plot the average value along with the maximum and minimum value.

\begin{figure}[ht]
\centering
\includegraphics[width=0.8\columnwidth]{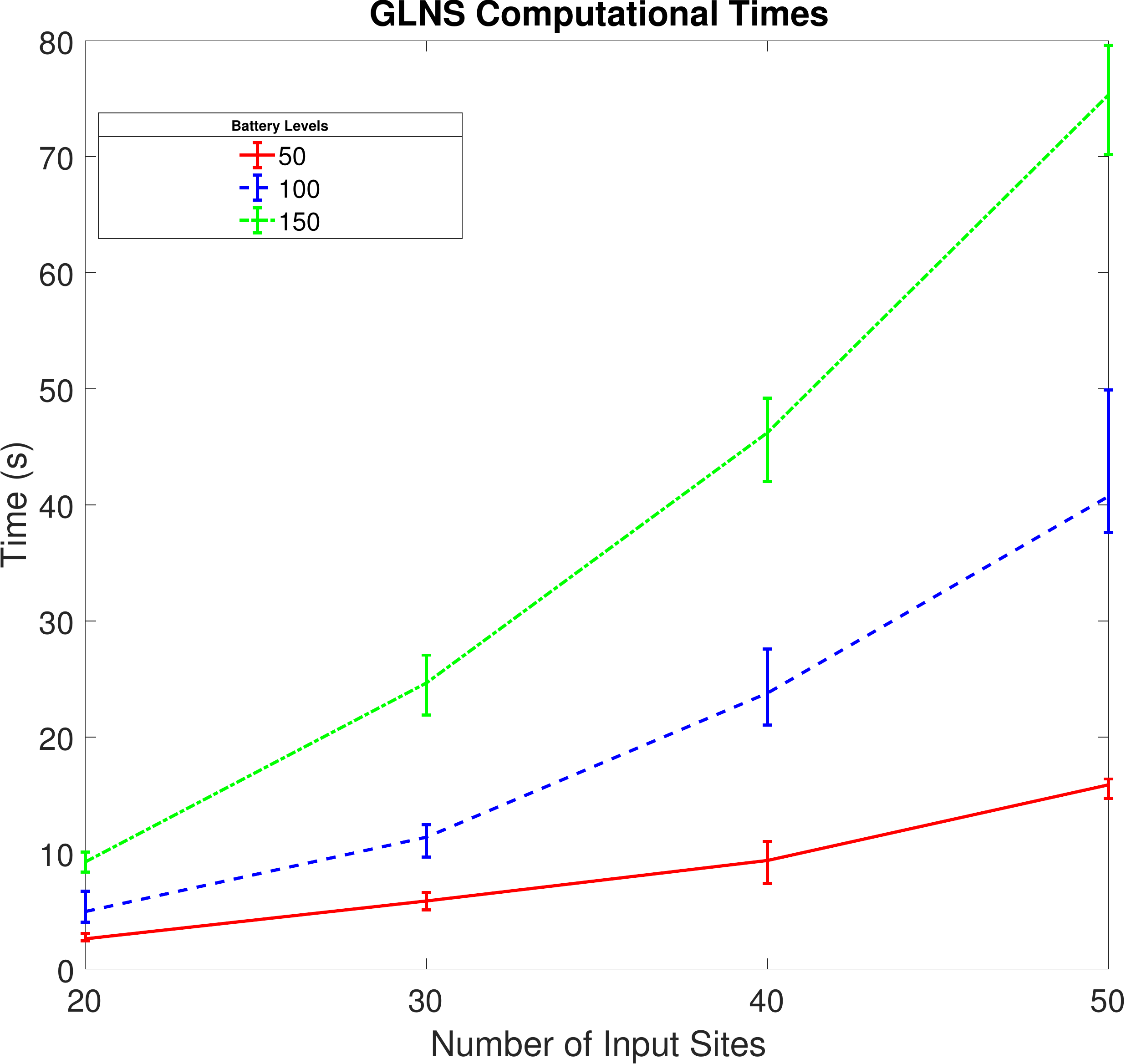}
\caption{The computational time of GLNS for larger instances.}
\label{fig:GLNS_compTime}
\end{figure}

Due to limitations in \emph{concorde}, we were not able to run larger instances, but GLNS can run larger instances, as shown in Figure \ref{fig:GLNS_compTime}. We show the effect of incrementing $n$ from 20 to 50 and $m$ from 50 to 150 in Figure \ref{fig:GLNS_compTime}. We ran 5 random instances and plot the average of the 5 instances with the maximum and minimum value.

\begin{figure}[ht]
\centering
\includegraphics[width=\columnwidth]{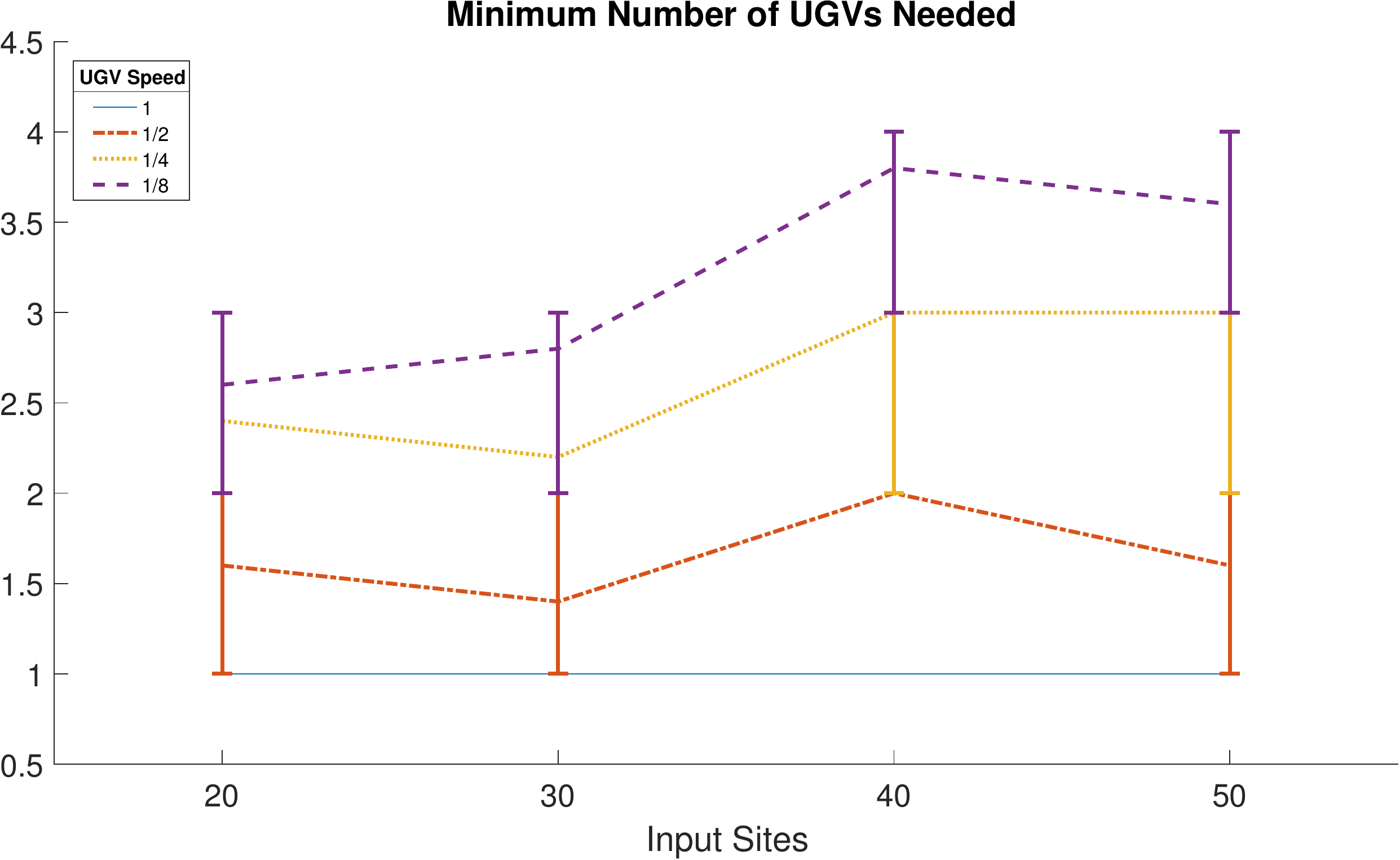}
\caption{Minimum number of UGVs necessary to service an UAV. The input was the $m = 50$ data from Figure \ref{fig:GLNS_compTime}.}
\label{fig:ugvmin}
\end{figure}

Figure \ref{fig:ugvmin} shows the minimum number of UGVs necessary to service a single UAV. We used the same $m = 50$ data set that was used to create Figure \ref{fig:GLNS_compTime}. The plot shows the effects of having a slower UGV.

\subsection{Preliminary Field Experiments}
\begin{figure}[ht]
\centering
\begin{subfigure}[b]{0.45\columnwidth}
\includegraphics[width=\textwidth, height = 5cm]{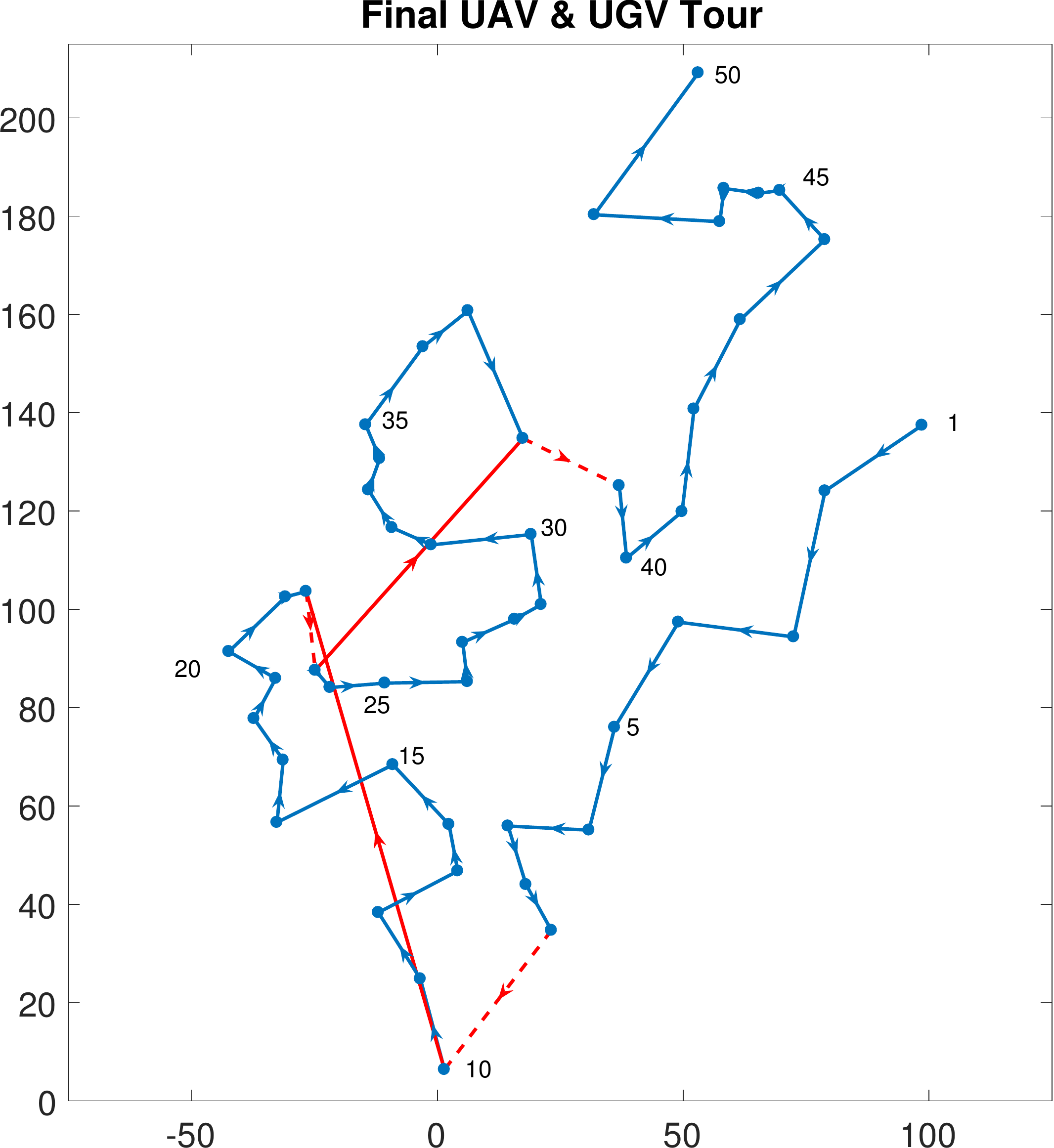}
\caption{Final flight path of the UAV and path of the UGV.}
\label{fig:kentInit}
\end{subfigure}
\begin{subfigure}[b]{0.45\columnwidth}
\includegraphics[width=\textwidth]{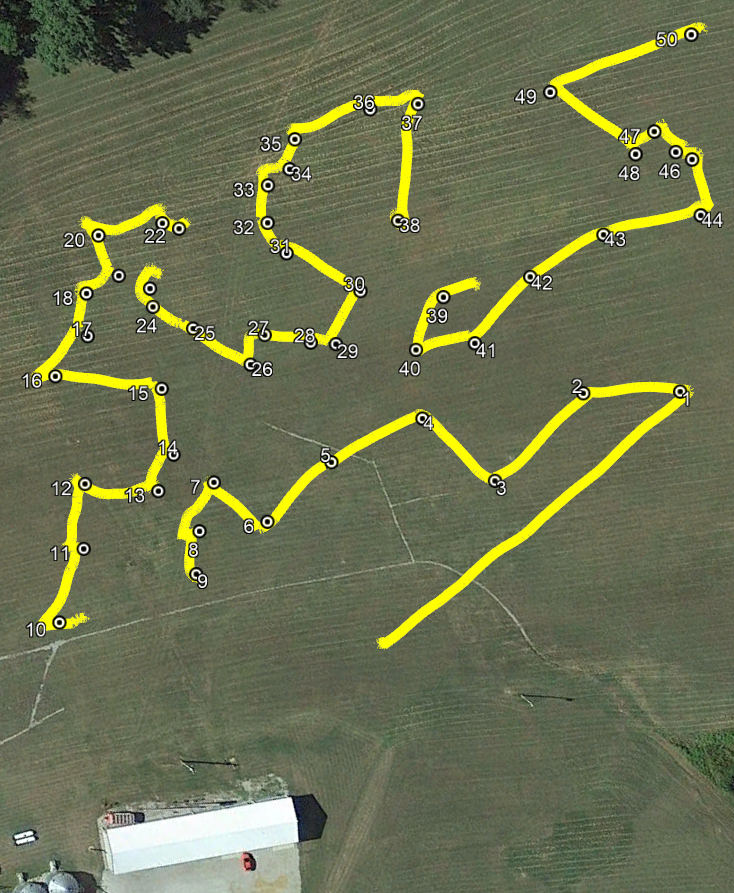}
\caption{GPS trace of the UAV.}
\label{fig:realWorld}
\end{subfigure}
\caption{Experiments were performed at Kentland Farms in Blacksburg VA. The input was 50 sites in a $200 \times 100$ meters area with 100 battery levels. $t_{TO} = 4$, $t_{L} = 4$, $r = 0$, and $t_{UGV} = t_{UAV}$. For the experiment the UAV was manually landed and the UGV was driven manually from landing to take off. All other UAV actions were done autonomously. More results are included in the multimedia submission.}
\label{fig:kent}
\end{figure}

We also carried out preliminary field experiments using the quadrotor and Husky UGV (Figure~\ref{fig:realUAV}). The input consisted of 50 sites shown in Figure \ref{fig:kentInit} and $m = 100$ battery levels. We restricted the size of the area to a $200 \times 100$ meters area and set $t_{TO} = 4$, $t_{L} = 4$, $r = 0$, and $T_{UGV} = T_{UAV}$. Figure~\ref{fig:kent} shows the results of the experiments. The outputs for the UAV tour and UGV tour are shown in Figure~\ref{fig:kentInit}. Figure~\ref{fig:realWorld} shows the output data from the Pixhawk flight controller. The UAV was programmed to fly autonomous GPS missions. The autonomous missions were executed by sending a set of waypoints that created a path for the UAV to travel and meet up with the UGV. Once at the final waypoint on the path, the we manually piloted the UAV to land on the UGV. The UGV was manually driven to the next take-off point in the tour. This next waypoint could be a new waypoint due to \typetwo edge or the same waypoint due to \typethree edge. The UAV then detects if it is at the next take-off point by comparing its current GPS position with the GPS coordinates of the next take-off point. This process loops until the UAV reaches its final waypoint whereupon the UAV would autonomously land on the ground. Additional experiments can be seen in the multimedia submission.

\section{Conclusion and Future work}
In this paper, we present an optimal algorithm for routing a battery-limited UAV and a mobile recharging station to visit a set of sites of interest. We are also conducting larger scale experiments using a recharging station being developed in-house. The longer-term future work is to design algorithms to handle multiple UAVs and UGVs as well as stochastic energy consumption models.

\bibliographystyle{plain}
\bibliography{ref}
\end{document}